\documentclass[12pt, onecolumn,draftcls]{IEEEtran}

\usepackage[numbers,sort&compress]{natbib}
\usepackage[caption=false,font=scriptsize]{subfig}
\usepackage{url}
\usepackage{color}
\usepackage{multirow}

\usepackage{amssymb,amsmath}

\usepackage{geometry}
\ifCLASSINFOpdf
   \usepackage[pdftex]{graphicx}
   \graphicspath{{./}}
\else
\fi
\begin{document}
\sloppy

%
\title{Energy Spatio-Temporal Pattern Prediction for Electric Vehicle Networks} 

\author{\IEEEauthorblockN{Qinglong Wang}
}



\maketitle

\begin{abstract}
Information about the spatio-temporal pattern of electricity energy carried by EVs, instead of EVs themselves, is crucial for EVs to establish more effective and intelligent interactions with the smart grid. In this paper, we propose a framework for predicting the amount of the electricity energy stored by a large number of EVs aggregated within different city-scale regions, based on spatio-temporal pattern of the electricity energy. The spatial pattern is modeled via using a neural network based spatial predictor, while the temporal pattern is captured via using a linear-chain conditional random field (CRF) based temporal predictor. Two predictors are fed with spatial and temporal features respectively, which are extracted based on real trajectories data recorded in Beijing. Furthermore, we combine both predictors to build the spatio-temporal predictor, by using an optimal combination coefficient which minimizes the normalized mean square error (NMSE) of the predictions. The prediction performance is evaluated based on extensive experiments covering both spatial and temporal predictions, and the improvement achieved by the combined spatio-temporal predictor. The experiment results show that the NMSE of the spatio-temporal predictor is maintained below $0.1$ for all investigate regions of Beijing. We further visualize the prediction and discuss the potential benefits can be brought to smart grid scheduling and EV charging by utilizing the proposed framework.
\end{abstract}

%
\IEEEpeerreviewmaketitle

\section{Introduction}\label{sec:intro}

Electric vehicles (EVs) \cite{CST} are increasingly regarded as not only a promising alternative to fossil fuel engine vehicles for environmental and efficiency concerns, but also flexible distributed energy facilities which are capable of both consuming and providing electricity energy via interacting with the smart grid \cite{Chunhua2013Opportunities,Wencong2012Survey,du2014distributed}. Recent thriving development of battery techniques has greatly accelerated this tendency \cite{Tuttle2012Evolution}. While batteries of EVs have been advanced in term of both capacity and power density, this in turn indicates that EVs will consume more energy with larger power. For example, according to the U.S. Energy Information Administration \cite{howmuchelectricity}, in 2013, a U.S. residential utility customer consumes an average of $909$kWh of electricity per month, while an EV with a $24$kWh battery pack consumes around $720$kWh if charged once daily. Meanwhile, the charging power can be as high as $19.2$kW when charged by level 2 charging standard ($240$V and $80$A) \cite{Yilmaz2013Review}. Obviously, electricity load from EV charging would pose significant impacts on the smart grid if not handled properly \cite{GreenII2011544,Clement2010Impact}. Meanwhile, EVs also bring new opportunities to the smart grid. Utilizing EVs to provide regulation services and developing Vehicle-to-Grid (V2G) technique have been more and more intensively discussed \cite{Chunhua2013Opportunities,Wencong2012Survey,Tuttle2012Evolution}.

Indeed, it is actually the electricity energy carried by EVs that can influence and also benefit the smart grid. As EVs moving across streets, districts and even cities, the electricity energy is carried along. Hence, the spatial pattern of EVs' mobility is inherited by the electricity energy stored in battery packs. Meanwhile, the stored energy keeps decreasing as an EV keeps driven by electric power. The changes of the stored energy also possess temporal pattern. Therefore, to mitigate the impact and also improve the efficiency of using EVs' batteries for providing regulation and V2G services, it is crucial to discover the spatio-temporal pattern of the electricity energy carried by EVs.

Spatially related energy patterns can be explored by observing the mobility pattern of EVs. There have been many research works aiming at studying traffic pattern \cite{Fabritiis2008Traffic,Yuan2010Yuan, Yuan2011Driving} and driving pattern of individual vehicle \cite{Qiuwei2010Driving,Lei2014GeoMob,Lee2012Stochastic}. However, spatial pattern of the energy carried by EVs are rarely explored but is critical for large-scale energy planning. Meanwhile, temporal changes of energy are due to the usage of electricity energy for propulsion. Research works studying the charging and discharging scheduling of EVs either focus on modeling energy changes of an individual EV \cite{Lee2012Stochastic, Donadee2014Stochastic}, or of aggregated EVs restricted in static scenarios, where EVs are plugged-in and the scheduling is only for the plugged-in period \cite{Yifeng2012Optimal,Ardakanian2014Real,Sortomme2011Coordinated}. \cite{Miao2014Mobility} has considered the influence of EVs' mobility on the stored energy and proposed a coordinated charging strategy.

Our work is different in that we view EVs' energy pattern through both spatial and temporal lens. We aim at building an energy spatio-temporal pattern prediction framework, upon which crucial questions including when, where and how much is the available energy for effective scheduling can be answered. To realize this framework, various types of sensing and monitoring facilities are needed. Although there have been many existing techniques available to capture and transmit EVs' mobility information, e.g. GPS, vehicular ad-hoc networks (VANETs) \cite{Yujin2013Horizon,Lei2014GeoMob, ToN}, dedicated short range communication (DSRC) \cite{Kenney2011Dedicated}, cellular networks \cite{Eltahir2014Vehicular, du2013network} and Wi-Fi \cite{Shaohan2014Towards, clockconf}, etc. However, when the focus is on energy pattern, these techniques are not sufficient. We need additional functionality of providing information of energy storage status, e.g. state-of-charge (SOC). This can be realized through transmitting SOC information from an EV to road-side communicating facilities. e.g. road-side-units (RSUs) \cite{Miao2014Mobility,Liang2012Double}, and further submitting SOC information from RSUs to central control facilities. Our work shows that, with communicating techniques and facilities that are capable of collecting SOC information, EVs' energy pattern can be discovered, thus contributes to establishing more intelligent interaction with the smart grid \cite{7037294}.

In summary, the contributions of this paper are as follows:
\begin{enumerate}
	\item We propose a spatio-temporal prediction framework to predict the aggregated energy carried by a large number of EVs within city-scale regions. Both spatial and temporal patterns are utilized for building a combined spatio-temporal predictor.
	\item We extract both spatial and temporal features for exploring the hidden correlation between them and the changing pattern of the aggregated energy.
	\item We utilize a real city-scale trajectories dataset for feature extraction and prediction performance evaluation. Low prediction errors demonstrate the effectiveness of the proposed spatio-temporal framework.
\end{enumerate}

The rest of this paper is organized as follows: Section II introduce the data set used and spatial and temporal features extraction procedures. In Section III we provide an overview of the energy spatio-temporal pattern prediction framework. Section IV presents the design of the spatial predictor and the temporal predictor in detail. We evaluate the prediction performance of the proposed predictor in Section. V, including the demonstration of the influence on the prediction performance by selecting different features, and the comparison of the prediction performance of the spatial predictor, the temporal predictor and the spatio-temporal predictor. Finally, Section VI concludes the paper and address our future work.

\section{Dataset and Feature Extraction}\label{sec::dataset}
\begin{figure}[t]
	\begin{center}
		\includegraphics[scale=0.33]{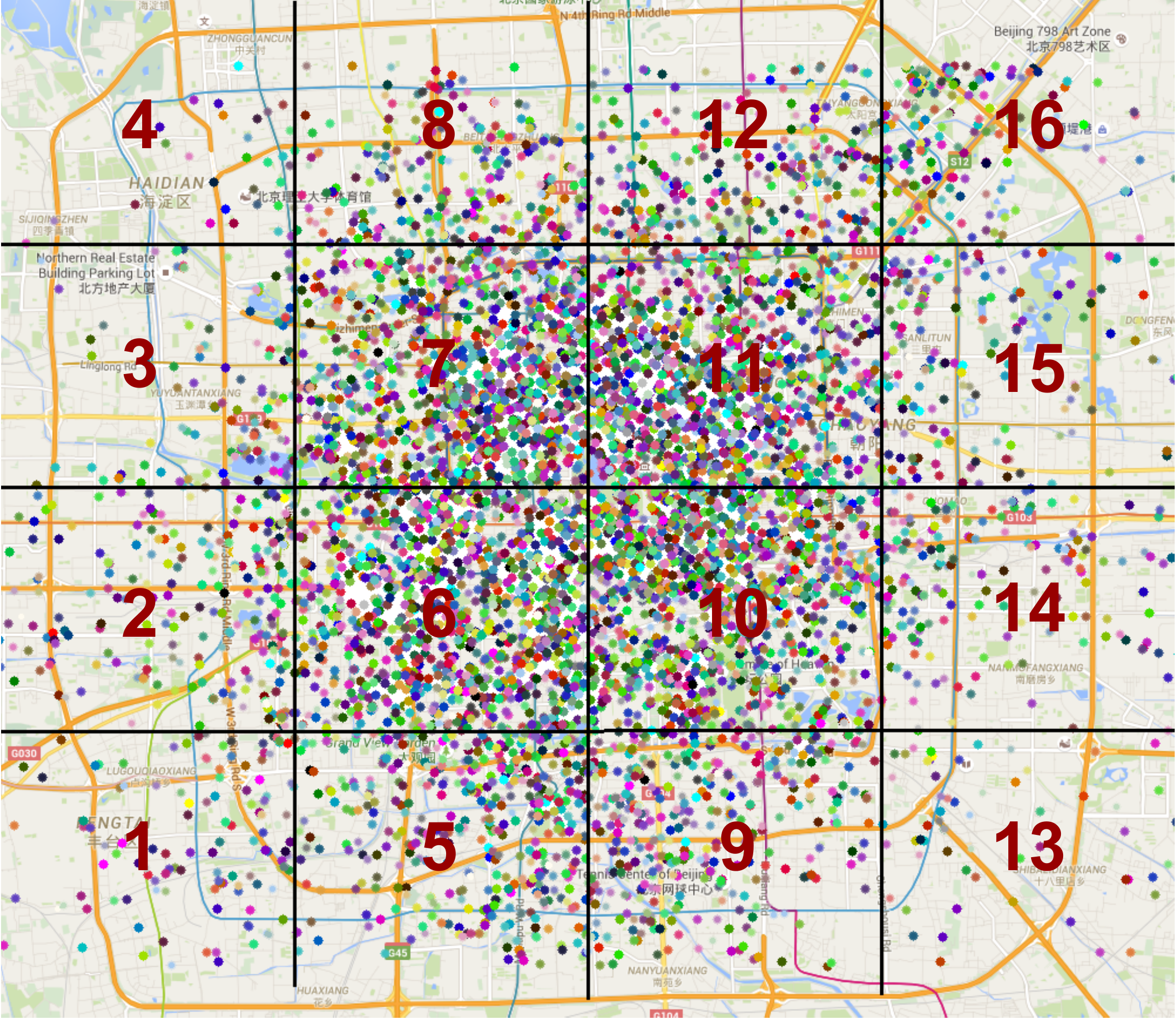}\\
		\caption{Spatial distribution of all recorded taxis over sixteen divided regions within 4th Ring Road of Beijing.}
		\label{fig::beijing4ring}
	\end{center}
\end{figure}
The dataset used in this paper is based on a GPS trajectories dataset of $10,357$ taxis from Feb. 3 to Feb. 8, 2008 in Beijing \cite{Zheng2011dataset}. As most of taxis recorded travel within the $4$th Ring Road of Beijing, we target on this area and further divide it into sixteen disjointed regions, as shown in Fig. \ref{fig::beijing4ring}. Within each region, we assume there exist at least one road-side facility responsible for collecting the SOC information reported by each EV, and transmitting it to the central controller.

Assuming all the trajectories data is generated by EVs, we further introduce the extraction of both spatial and temporal features based on the trajectories data. In term of spatial features, for each divided region, information related to the quantity, velocities and moving directions of EVs within such a region is considered. In term of temporal features, the hourly aggregated energy of all EVs within each region and the associated recording time is included. Table. \ref{tab::feature_set} shows the extracted spatial and temporal features for the $k$th region, $k = 1,\cdots,16$.
\begin{table}[t]
	\centering
	\caption{All extracted features for region $k = 1,\cdots,16$}
	\label{tab::feature_set}
	\begin{tabular}{|c|c|l|c|}
		\hline
		Features                          & \multicolumn{2}{c|}{Components}    & Meaning                                                                       \\ \hline
		\multirow{4}{*}{Spatial features} & \multicolumn{2}{c|}{$F_{N}^{k}$}& Number of EVs                                                                                                 \\ \cline{2-4}
		& \multicolumn{2}{c|}{$F_{V}^{k}$}& \begin{tabular}[c]{@{}c@{}}Average velocity: $V_{ave}^{k}$\\ Velocity variance: $V_{var}^{k}$\end{tabular}      \\ \cline{2-4}
		& \multicolumn{2}{c|}{$F_{D}^{k}$}& \begin{tabular}[c]{@{}c@{}}Number of EVs in direction 1: $D_{1}^{k}$\\ Number of EVs in direction 2: $D_{2}^{k}$\\ Number of EVs in direction 3: $D_{3}^{k}$\\ Number of EVs in direction 4: $D_{4}^{k}$\\ Variance of the number of EVs\\ in four directions : $D_{var}^{k}$\end{tabular} \\ \cline{2-4}
		& \multicolumn{2}{c|}{$F_{E}^{k}$}      & \multicolumn{1}{l|}{\begin{tabular}[c]{@{}c@{}}Aggregated energy: $E_{sum}^{k}$\\ Aggregated energy variance: $E_{var}^{k}$\end{tabular}}                                     \\ \hline
		\multirow{2}{*}{Temporal features}& \multicolumn{2}{c|}{$F_{E}^{k}$}      & Aggregated energy: $E_{sum}^{k}$ \\ \cline{2-4}
		& \multicolumn{2}{c|}{$F_{H}^{k}$}         & Recording hour: H                                                                                     \\ \hline
	\end{tabular}
\end{table}
\subsection{Spatial Feature Extraction}
\label{sec::spatial_features}

There have been many models proposed for studying traffic patterns. For example, \cite{Helge1995Mathematical,lammer2008self} draw an analogy between circuit and traffic networks, and use Kirchhoff's law to constrain traffic network models. Building models of this type can be effective for a single road segment or an intersect, as the traffic flow is constrained by explicit and simple conditions. However, if viewed at a much larger scale (for example, the whole city as studied in this paper), models of this type \cite{AsynNetwork} can bring much complexity when considering the countless roads, lanes and intersects, etc. Moreover, as we aim to explore the hidden correlation between the aggregated energy, instead of the traffic, among adjacent areas, simply employing any specific models may lead to severe bias problems.
\begin{figure*}[t]
	\begin{center}
		\includegraphics[scale=0.32]{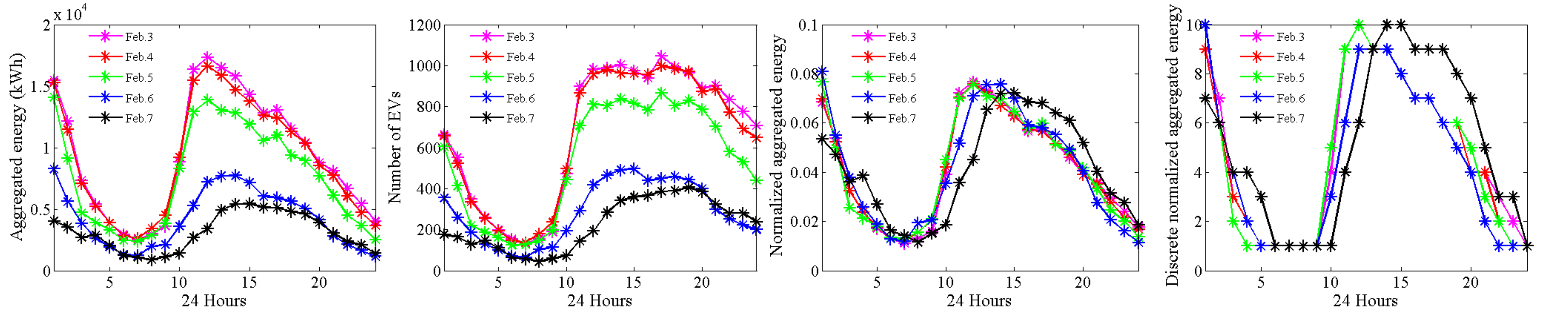}\\
		\caption{Demonstration of energy temporal patterns of a selected region. From left to right are aggregated energy, recorded number of EVs, normalized aggregated energy and discrete normalized aggregated energy of $24$ hours of five days. Especially, in second graph on the left, the recorded numbers of EVs of all observed days share a similar varying pattern. This implies that most taxis maintain similar daily driving pattern, which can be applicable for private EVs.}
		\label{fig::temporal_fea}
	\end{center}
\end{figure*}

Up to the present, there is no clear form for modeling the flow patterns of energy when having moving vehicles as the carriers. However, luckily there have been many research works employing neural networks to handle the complex modeling of traffic networks \cite{Dia2001253,zheng2006short}. Therefore, as shown in Table. \ref{tab::feature_set}, we extract dynamic spatial features representing general aspects of a selected area (compared with specific ones representing a single road or intersect). Meanwhile, various static features including terrain, real estates along road segments, road number and capacity have been proposed in \cite{Zheng2013UAir}. Note that although these static features also represent a specific region, they remain unchanged and can hardly represent the changing pattern of energy.

As shown in Table. \ref{tab::feature_set}, for each region $k = 1,\cdots,16$, spatial features contain following four feature sets:
\begin{itemize}
	\item Firstly, we use $F_{N}^{k}(t)$ to denote the number of EVs within region $k$ during $t$th recording hour. $F_{N}^{k}(t)$ is calculated based on the GPS location and timing records contained in the original trajectories dataset.
	\item Secondly, we calculate the driving distance $S_{m-1,m}^{n}$ between two adjacent GPS locations $P_{m-1}^{n}$ and $P_{m}^{n}$ for EV $n$. We also calculate the time difference $\Delta t_{m-1,m}^{n}$ between corresponding time records. Hence, the velocity of EV $n$ at recording time $t$ can be obtained as $V_{m-1,m}^{n}(t+1) = S_{m-1,m}^{n}/\Delta t_{m-1,m}^{n}$. Furthermore, we get additional features including the average and variance of the velocity as $V_{ave}^{k}(t)$ and $V_{var}^{k}(t)$ for region $k$.
	\item Thirdly, we utilize GPS location records to calculate the moving direction $\mathbf{d}_{m}^{n}$ of EV $n$, as $\mathbf{d}_{m}^{n} = \left [P_{m}^{n}, P_{m-1}^{n}\right ]^{\mathrm{ T }}$. $\mathbf{d}_{m}^{n}$ is further determined to be within one of four quadrants $D_{1}$, $D_{2}$, $D_{3}$ and $D_{4}$. With $\mathbf{d}_{m}$ of all EVs, we calculate the variance of the moving directions as $D_{var}^{k}$ for region $k$.
	\item Fourthly, we obtain the energy related feature set $F_{E}^{k}$ and recording time $F_{H}^{k}$ as introduced in Section. \ref{sec::temp_fea}.
\end{itemize}

After calculating the spatial features of each region, these features are fed to a neural network based spatial predictor to capture the hidden correlation of the aggregated energy among adjacent regions. The input for the spatial predictor consists of the four spatial feature sets, $F_{N}$, $F_{V}$, $F_{D}$ and $F_{E}$ of all neighbours of region $k$ (For example, for region $6$, its neighbours include region $1$, $2$, $3$, $5$, $7$, $9$, $10$ and $11$. While for regions residing at the margins of Fig. \ref{fig::beijing4ring}, take region $1$ as an example, we pick regions $2$, $3$, $5$, $6$, $7$, $9$, $10$ and $11$ as its neighbours).

\subsection{Temporal Feature Extraction}
\label{sec::temp_fea}
Temporal features consists of recording hour $H_{h}$, where $h = 1,...,24$ and energy related feature set $F_{E}$. For the former, it can be easily obtained by checking the GPS timing records. The latter is extracted based on previously calculated distance $S_{m-1,m}^{n}$ of EV $n$. We first assume all the EVs have the same battery capacity $C = 24kWh$. Then we extract the SOC of each EV by setting it linearly dependent on the distance this EV is driven, i.e. the more distance an EV travels, the lower is its SOC. This simplified model has also been used in \cite{Hao2013Optimizing,Miao2014Mobility}.

Since the trajectories data is generated by taxis, we notice that most taxis recorded keep moving even at midnight. This implies that there is no universal shift time for these taxis. Therefore, we need to assign an universal recharging time for all these taxis, given we have assumed them to be EVs. We assume they are all recharged to full capacity (i.e. SOC is initiated to $1$) at $0$ a.m. of every observed day from Feb. 3 to Feb. 8, 2008. This assumption is practical when trajectories data of common private EVs is available, since people usually have their EVs recharged to full capacity for next day's drive. In this case, SOC of private EVs are most likely to be initiated to $1$ before an universal work time.

Similar to spatial feature extraction in Section. \ref{sec::spatial_features}, we calculate the variance of remaining electricity $E_{var}^{k}(t)$ (obtained via using all EVs' reported SOC and capacity $C$) and the aggregated remaining energy $E_{sum}^{k}(t)$ of all EV within $k$th region at time $t$. The hourly updated $E_{sum}^{k}(t)$ of six observed days of an example region is shown in the first graph on the left in Fig. \ref{fig::temporal_fea}. From Fig. \ref{fig::temporal_fea}, we notice that all observed days of this region share a similar varying pattern of aggregated energy, although with different scales. This can be explained by observing the fluctuation of the number of EVs in this region (shown in the second graph on the left in Fig. \ref{fig::temporal_fea}). For all observed days, EVs crowd into and depart from this region at similar time. Take 12 p.m. of every day as an example, when the aggregated energy starts to drop, the number of EVs remains at the maximal value for hours. This is because when the number of EVs first reaches the highest value, EVs have more remaining energy, compared with hours later when the number of EVs starts to drop. The peak of aggregated energy lasts shorter than that of the number of aggregated EVs.

As it is the varying pattern of aggregated energy that we aim to capture, it is necessary to mitigate the impact of the scale differences of the aggregated energy among each observed days. We normalize the aggregated energy $E_{sum}^{k}(t)$ at time $t$ by the summation $\sum_{t=1}^{24}E_{sum}^{k}(t)$ of the day it is recorded. The normalized result is shown in the third graph on the left in Fig. \ref{fig::temporal_fea}. Clearly, the varying pattern of aggregated energy is well preserved.

Furthermore, in general, continuous variable CRF \cite{informationmatrix, asilomar} often leads to exponential computation complexity \cite{lafferty2001conditional, pairwise}. In this paper, we adopt discrete variable CRF. We discretize the normalized aggregated energy into $10$ levels, which is sufficient to represent the varying patterns. The discretized aggregated energy is shown in the fourth graph on the left in Fig. \ref{fig::temporal_fea}.



\section{Framework Overview}\label{sec::architecture}
\begin{figure}[t]
	\begin{center}
		\includegraphics[scale=0.4]{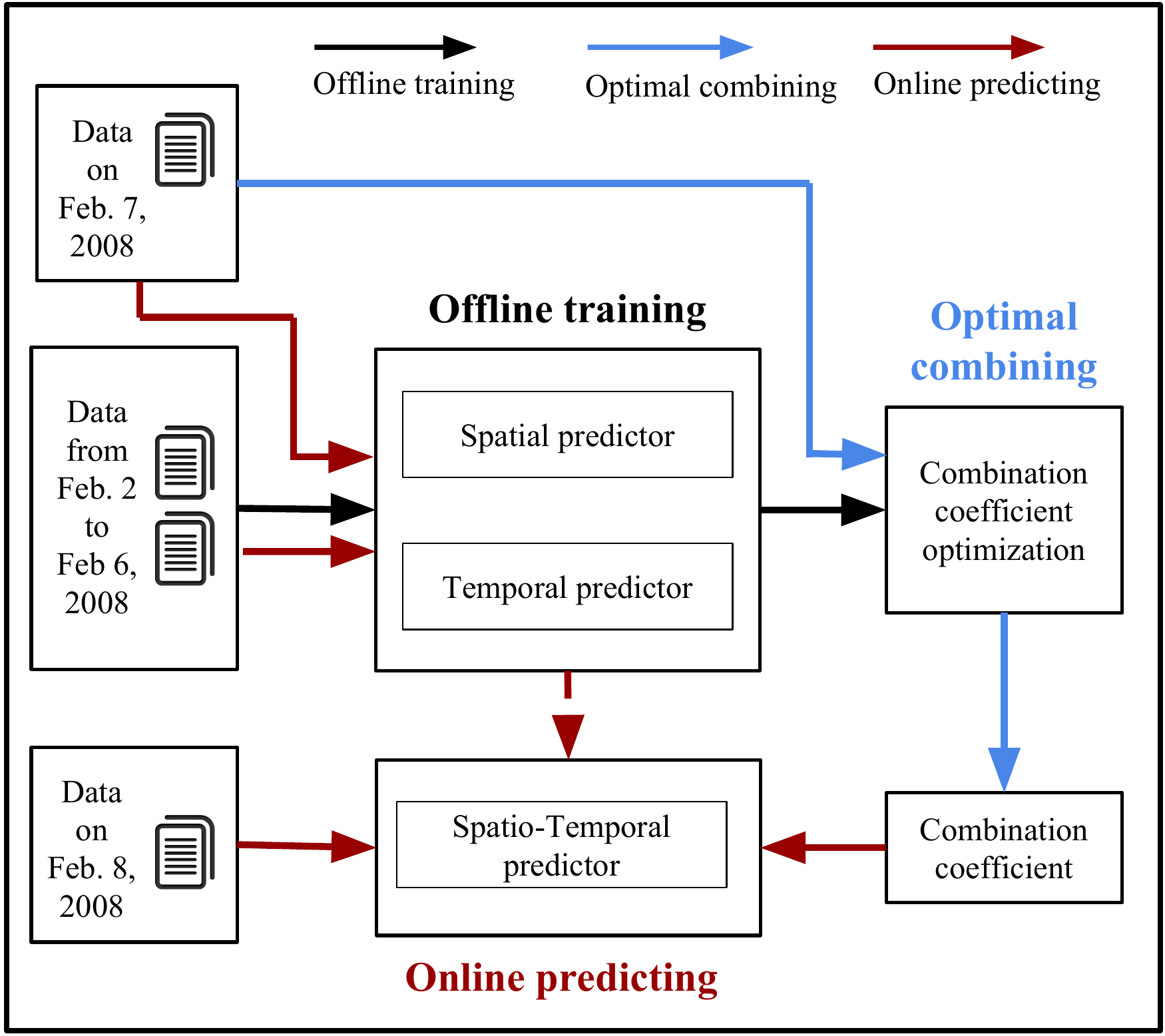}\\
		\caption{Overview of energy spatio-temporal pattern prediction framework.}
		\label{fig::frame}
	\end{center}
\end{figure}

The energy spatio-temporal prediction framework consists of three phases: (1) offline training of predictors, (2) optimal combining of both predictors and (3) online predicting, as shown in Fig. \ref{fig::frame}. We split both temporal and spatial features into a training set, a validation set and a testing set respectively. During offline training, the spatial and temporal predictors are trained using the training set independently. In the second phase, two predictors are combined via using a combination coefficient $\lambda$. The optimal $\lambda^{*}$ is obtained by minimizing the normalized mean square error (NMSE) of the validation set. Finally, both trained predictors and the optimal combination coefficient $\lambda^{*}$ are tested by the testing set.
\subsection{Offline Training}
During offline training, we first select temporal and spatial features from Feb. 3 to Feb. 6, 2008 as the training set for the temporal predictor and the spatial predictor respectively.
\subsubsection{Training Temporal Predictor}
The linear-chain CRF based temporal predictor is fed with temporal features (i.e. recording hour $F_{H}^{k}$ and energy related features $F_{E}^{k}$) of each region $k = 1,...,16$, and then trained independently.
\subsubsection{Training Spatial Predictor}
Spatial features (i.e. $F_{N}^{k}$, $F_{V}^{k}$ and $F_{D}^{k}$ of all neighbour regions of region $k$) are gathered independently for each region. The training for the spatial predictor is hourly based, meaning that the spatial predictor takes all spatial features at current time $t$ as input, while taking the actual aggregated energy value at time $t+1$ as target.
\subsection{Optimal Combining}
We take temporal and spatial features on Feb. 7, 2008 as the validation set for obtaining the optimal combination coefficient $\lambda^{*}$. Both temporal and spatial features are fed into the trained temporal and spatial predictor respectively. Since the SOC of all EVs are initiated to $1$ at the beginning hour of each day (see Section. \ref{sec::dataset} for detail), we need to assign an initial value for the temporal predictor to start. We average the aggregated energy values of the starting hour of the entire training set as the initial value. The predictions of both temporal predictor and spatial predictor are conducted hourly and independently. The temporal prediction results are the hourly probability distributions of different levels of aggregated energy for each region, while the spatial prediction results are the aggregated energy values of all regions. After obtaining the prediction results of all $24$ hours, the temporal and spatial prediction results for all regions are combined via $\lambda$, for which the optimal $\lambda^{*}$ minimizes the NMSE of predictions for the validation set.
\subsection{Online Predicting}
Online predicting is similar to the procedure of optimizing combination coefficient, except that the optimal $\lambda^{*}$ has been set. Both spatial predictor and temporal predictor are retrained with features extracted from Feb. 2 to Feb. 7, 2008, and tested by features extracted on Feb. 8, 2008. Additionally, during online predicting, spatial predictors are retained repeatedly every hour when new observations are available. For temporal predictor, it is retrained when the observations of an entire new day are obtained.

\section{Spatial \& Temporal Learning}
In this section, we introduce the design of the spatial predictor and the temporal predictor in detail. Then the combining method for building the spatio-temporal predictor is presented. The optimal combining coefficient is obtained through minimizing the NMSE of the prediction results generated by the spatio-temporal predictor.
\subsection{Design of the Spatial Predictor}
We utilize neural network to capture the complex spatially related energy pattern, which inherits the changing pattern of moving traffic. For the latter, neural network has been widely employed for performing traffic predictions \cite{Dia2001253,zheng2006short}. We utilize a two-layer feedforward neural network, of which the neurons in the hidden layer are associated with a sigmoid transfer function, and a linear function for the output layer.

For each region $k$, its spatial features are $\mathbf{I}^{k} = \left \{\mathbf{F}_{N}^{k}, \mathbf{F}_{V}^{k}, \mathbf{F}_{D}^{k}, \mathbf{F}_{E}^{k}\right \}^{\mathrm{T}}$, of which each component denotes a sequence of hourly recorded spatial features of region $k$.
For example, $\mathbf{F}_{N}^{k} = \left [ F_{N}^{k}(1),\cdots , F_{N}^{k}(T_{train})\right ]^{\mathrm{T}}$, where $T_{train}$ is the total number of time slots used for training. The input to the neural network for region $k$ is $\mathbb{I}^{k} = \left \{\mathbf{I}^{k_{1}},\cdots,\mathbf{I}^{k_{N}}\right \}$, which is the set of spatial features of all its neighbor regions $k_{1},\cdots,k_{N}\in \mathfrak{R}_{k}$, where $\mathfrak{R}_{k}$ denotes the neighbour set of region $k$. Meanwhile, the target is $\mathbb{O}^{k} = \mathbf{E}^{k}_{sum}$, where $\mathbf{E}^{k}_{sum}$ is a sequence of aggregated energy within the $k$th region, i.e. $\mathbf{E}^{k}_{sum} = \left [ E^{k}_{sum}(1+\Delta T),\cdots, E^{k}_{sum}(T_{train} + \Delta T) \right ]^{\mathrm{T}}$. $\Delta T$ is the prediction time interval, which can be set from $1$ to $24$.

The testing of the spatial predictor is conducted hourly via using newly measured features at time $t$ as input to predict the aggregated energy value in the following time slot $t+1$. When the measured features of time $t+1$ are available, these newly obtained features are appended to the training set, with which the spatial predictor is retrained.
\subsection{Design of the Temporal Predictor}
We employ a linear-chain CRF to build our temporal predictor, which utilizes temporal features of each region to predict the future aggregated energy within that region. A linear-chain CRF has the advantages of relaxing independence assumption over hidden Markov models and avoiding the fundamental label bias problem over maximum entropy Markov models and other discriminative Markov models based on directed graphical models \cite{lafferty2001conditional, duJMLR}.
\begin{figure}[b]
	\begin{center}
		\includegraphics[scale=0.4]{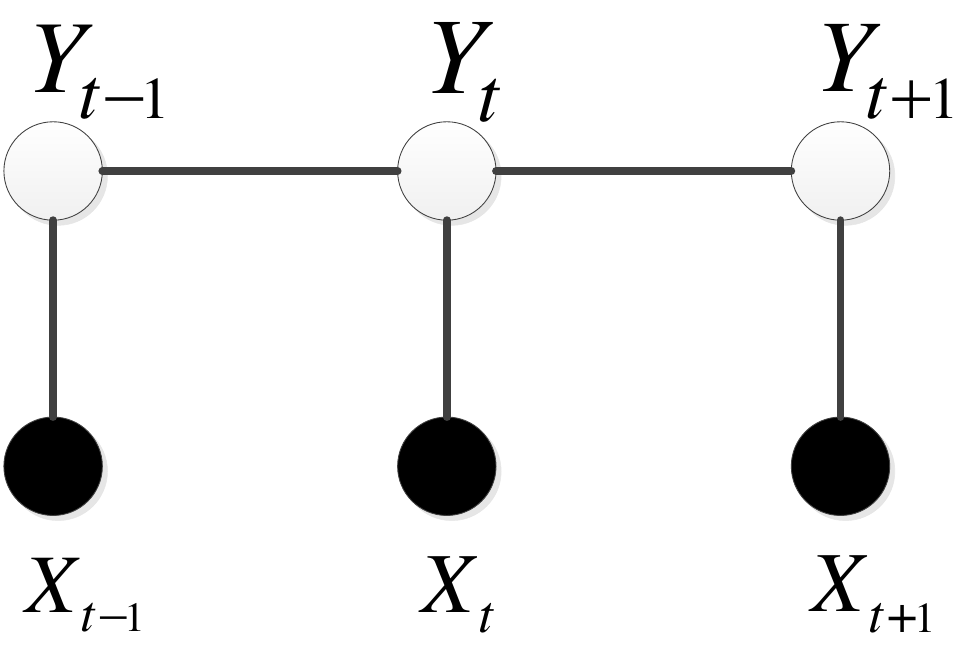}\\
		\caption{Graphic structure of a linear-chain CRF.}
		\label{fig::crf_model}
	\end{center}
\end{figure}

A linear-chain CRF is for building undirected probabilistic models for segmenting sequence data, as the graph $G = (V,E)$ shown in Fig. \ref{fig::crf_model}. $\mathbf{X}$ is the observation set, consisting of a sequence of observations $X_{t}$, i.e. $\mathbf{X} = \left \{X_{1}, X_{2},\cdots X_{T} \right \}$. In our case, we set each observation $X_{t}$ as the recording hour $F_{H}^{k}(t)$ of $k$th region at time $t$. Random state variable $\mathbf{Y}$ is indexed by the vertices of $G$, and consists of components $Y_{t}$, which varies over the label set comprising $10$ discrete levels of normalized aggregated energy. Given $\mathbf{X}$ and $\mathbf{Y}$, ($\mathbf{X},\,\mathbf{Y}$) is a CRF with a graph structure as shown in Fig. \ref{fig::crf_model}. Therefore, according to the Markov property, the conditional distribution of random state variables $Y_{t}$ follows:
\begin{equation}
\begin{aligned}
p(Y_{t}\mid \mathbf{X}, Y_{\tau}, \tau \neq t) = p(Y_{t}\mid \mathbf{X}, Y_{\tau}, \tau \sim t),
\end{aligned}
\end{equation}
where $\tau \sim t$ means that $\tau$ and $t$ are neighbors in graph $\mathbf{G}$. Note that the linear-chain CRF is globally conditioned on observation set $\mathbf{X}$.

Denote an observation sequence as $\mathbf{x}$ and a label sequence as $\mathbf{y}$ respectively, and have $\mathbf{y}\mid_{e}$ and $\mathbf{y}\mid_{v}$ represent the set of components of $\mathbf{y}$ associated with the edges and vertices of graph $G = (V,E)$. The conditional distribution of $\mathbf{y}$ given $\mathbf{x}$ is as follows:
\begin{equation}
\begin{aligned}
&p_{\theta}(\mathbf{y}\mid \mathbf{x}) \propto\\
& \quad \mathrm{exp} \Big ( \sum_{e\in E,\,t}\gamma_{t}f_{t}(e,\mathbf{y}\mid_{e},\mathbf{x}) + \sum_{v\in V,\,t} \mu_{t}g_{t}(v,\mathbf{y}\mid_{v},\mathbf{x})\Big),
\label{eqa::con_prob}
\end{aligned}
\end{equation}
where $f_{t}$ and $g_{t}$ are the edge and vertex potential functions which represent the features associated with edges and vertices respectively. $\theta = (\gamma_{1},\gamma_{2},\cdots ; \mu_{1},\mu_{2},\cdots)$ are numerical weights assigned to all features. In our case, we set both $f_{t}$ and $g_{t}$ as Boolean variables to indicates recording hour $F_{H}^{k}(t)$. Given training data $\mathcal{D} = \left \{\mathbf{x}_{n}, \mathbf{y}_{n}  \right \}_{n=1}^{N}$, where each $\mathbf{x}_{n}$ is a sequence of observation, and each $\mathbf{y}_{t}$ is a sequence of state variables, parameters $\theta$ can be estimated via maximizing the following conditional log-likelihood function:
\begin{equation}
\begin{aligned}
\mathcal{L}(\theta) = \sum_{t=n}^{N}\mathrm{log}\,p(\mathbf{y}_{n}\mid \mathbf{x}_{n}).
\label{eqa::likelihood}
\end{aligned}
\end{equation}

To combat over-fitting, we further use $L_{2}$ regularization, which penalizes vectors with too large norms. The maximization of (\ref{eqa::likelihood}) over $\theta$ is solved via gradient descent. Additionally, the computation of gradient requires marginal distribution for each edge transition, which is solved by forward-backward recursions.
\subsection{Optimal Combining Method}
After training the temporal predictor and spatial predictor, these two predictors are fed with a validation set that does not overlap with the training set already used.
For the temporal predictor, its predicted result is the probability distribution of the normalized aggregated energy over $10$ discrete levels. For example, we use $p_{1}^{k},\cdots,p_{10}^{k}$ to denote the probabilities for level $1,\cdots,10$ of region $k$ respectively. Meanwhile, each class $1,\cdots,10$ is associated with $Y_{1}^{k},\cdots,Y_{10}^{k}$ respectively, which is recovered via averaging previous observations (see Section. \ref{sec::architecture} for detail).
For the spatial predictor, the predicted result of time $t$ is $\hat{E}_{SP}^{k}(t)$, which is further associated with a combination coefficient $\lambda$. In this way, we combine the prediction results from both predictors at time $t$ as follows:
\begin{equation}
\begin{aligned}
&\sum_{i=1}^{10}\frac{p_{i}^{k}(t)}{\sum_{m=1}^{10}p_{m}^{k}(t)+\lambda}Y_{i}^{k}(t) + \frac{\lambda}{\sum_{m=1}^{10}p_{m}^{k}(t)+\lambda}\hat{E}_{SP}^{k}(t),\\
&= \sum_{i=1}^{10}\frac{p_{i}^{k}(t)}{1+\lambda}Y_{i}^{k}(t) + \frac{\lambda}{1+\lambda}\hat{E}_{SP}^{k}(t),\\
&= \frac{1}{1+\lambda}Y_{TP}^{k}(t) + \frac{\lambda}{1+\lambda}\hat{E}_{SP}^{k}(t),
\label{eqa::combine}
\end{aligned}
\end{equation}
where $Y_{TP}^{k}(t) = \sum_{i=1}^{10}p_{i}^{k}(t)Y_{i}^{k}(t)$.

While (\ref{eqa::combine}) is only for region $k$, we further combine the prediction results of all regions. An universal optimal $\lambda^{*}$ is obtained through minimizing the NMSE of predictions, as shown in (\ref{eq::min_mse}):
\begin{equation}
\begin{aligned}
& \underset{\lambda}{\text{min}}
& \sum_{k=1}^{K}\frac{\left \| \mathbf{E}^{k} - (\frac{1}{1+\lambda}\mathbf{Y}_{TP}^{k}+\frac{\lambda}{1+\lambda}
	\hat{\mathbf{E}}_{SP}^{k}) \right \|_{2}^{2}}{\left \| \mathbf{E}^{k} \right \|_{2}^{2}},
\end{aligned}
\label{eq::min_mse}
\end{equation}
where $\mathbf{E}^{k} = \left [ E^{k}(1),\cdots,E^{k}(T) \right ]^{\mathrm{T}}$, $\mathbf{Y}_{TP}^{k} = \left [ Y_{TP}^{k}(1),\cdots,Y_{TP}^{k}(T) \right ]^{\mathrm{T}}$, $\hat{\mathbf{E}}_{SP}^{k} = \left [ \hat{E}_{SP}^{k}(1),\cdots,\hat{E}_{SP}^{k}(T) \right ]^{\mathrm{T}}$.

\section{Experiments}
In this section, we first conduct experiments to show the influence on the prediction performance of using different features. Then we fix the feature sets to which generate the best prediction performance, and demonstrate the performance of the temporal and spatial predictor respectively, as well as the performance gain obtained by combining both predictors with the optimal combination coefficient. We further show the prediction performance changes when the prediction interval increases to more than one hour, to explore the ability of the spatio-temporal predictor for performing longer term prediction. At last, we provide a direct visualization of the prediction.
\subsection{Prediction Performance of Using Different Feature Sets}
We have extracted four categories of features (i.e. $F_{V}$, $F_{D}$, $F_{N}$ and $F_{E}$) as input for the spatial predictor, as shown in Table. \ref{tab::feature_set} in Section \ref{sec::dataset}. Among these, $F_{V}$, $F_{D}$ and $F_{N}$ are more closely related to EVs' mobility pattern. To elaborate their influence on the prediction performance, we first divide the whole feature sets into several combinations, and show the prediction performance of the spatial predictor and the spatio-temporal predictor based on these combinations respectively (the prediction performance of the temporal predictor is not shown since it utilizes only temporal features, which do not change in this experiment). Due to space limit, we use abbreviations SP, TP and STP for spatial predictor, temporal predictor and spatio-temporal predictor in following tables and figures.
\begin{table}[b]
	\centering
	\caption{Prediction performance based on different feature sets.}
	\label{tab::fea_perf}
	\begin{tabular}{|c|c|c|c|}
		\hline
		Feature sets (SP)                     & $\textrm{NMSE}_{\textrm{SP}}^{ave}$                             & $\textrm{NMSE}_{\textrm{SP}}^{min}$                          & $\textrm{NMSE}_{\textrm{SP}}^{max}$ \\ \hline
		$F_{V}$, $F_{N}$, $F_{E}$             & $0.2311$                                           & $0.9322$                                   & $0.0305$\\
		$F_{D}$, $F_{V}$, $F_{N}$, $F_{E}$    & $0.1585$                                           & $0.3541$                                   & $0.0367$ \\
		$F_{E}$, $F_{N}$                      & $0.1570$                                           & $0.3215$                                   & $0.0653$  \\
		$F_{N}$                               & $0.1450$                                           & $0.3588$                                   & $0.0615$\\
		$F_{E}$                               & $0.1247$                                           & $0.2593$                                   & $0.0359$ \\
		$F_{D}$, $F_{N}$, $F_{E}$             & $0.1185$                                           & $0.1851$                                   & $0.0649$\\ \hline
		Feature sets (STP)                    & $\textrm{NMSE}_{\textrm{STP}}^{ave}$                            & $\textrm{NMSE}_{\textrm{STP}}^{min}$                         & $\textrm{NMSE}_{\textrm{STP}}^{max}$ \\ \hline
		$F_{V}$, $F_{N}$, $F_{E}$             & $0.0800$                                           & $0.3265$                                   & $0.0124$ \\
		$F_{D}$, $F_{V}$, $F_{N}$, $F_{E}$    & $0.0558$                                           & $0.1310$                                   & $0.0211$ \\
		$F_{E}$, $F_{N}$                      & $0.0675$                                           & $0.1274$                                   & $0.0369$ \\
		$F_{N}$                               & $0.0446$                                           & $0.0825$                                   & $0.0186$ \\
		$F_{E}$                               & $0.0459$                                           & $0.0897$                                   & $0.0171$ \\
		$F_{D}$, $F_{N}$, $F_{E}$             & $0.0485$                                           & $0.0703$                                   & $0.0254$ \\ \hline
	\end{tabular}
\end{table}

In Table. \ref{tab::fea_perf}, different feature combinations are permutated in the descending order according to the prediction performance (NMSE) achieved based on them. As shown in Table. \ref{tab::fea_perf}, not all features contribute to improving prediction performance equally. Simply using all features does not lead to the best performance (but the second worst performance in this experiment). The worst performance is obtained by using all feature sets except $F_{D}$, while the best performance is obtained when the whole feature sets are used except $F_{V}$. This result indicates that moving direction related features are more important for achieving accurate prediction compared with velocity related features. This may because $F_{D}$ features can represent the interaction between each region and their neighbours, as they show directly the number of EVs moving between adjacent regions. Therefore, $F_{D}$ features are indirectly related to the energy flow among these regions. As for $F_{V}$, since the SOC of an EV is assumed to be linearly dependent on its traveling distance, velocity features $F_{V}$ have little correlation with energy varying pattern. Furthermore, it is also noticed that using either $F_{N}$ or $F_{E}$ solely cannot lead to the best prediction performance. This indicates that traffic related spatial features indeed help improve the prediction performance. In following experiments, we select the feature combination of $F_{D}$, $F_{N}$, $F_{E}$ due to their lowest prediction NMSE.
\begin{figure}[b]
	\begin{center}
		\includegraphics[scale=0.37]{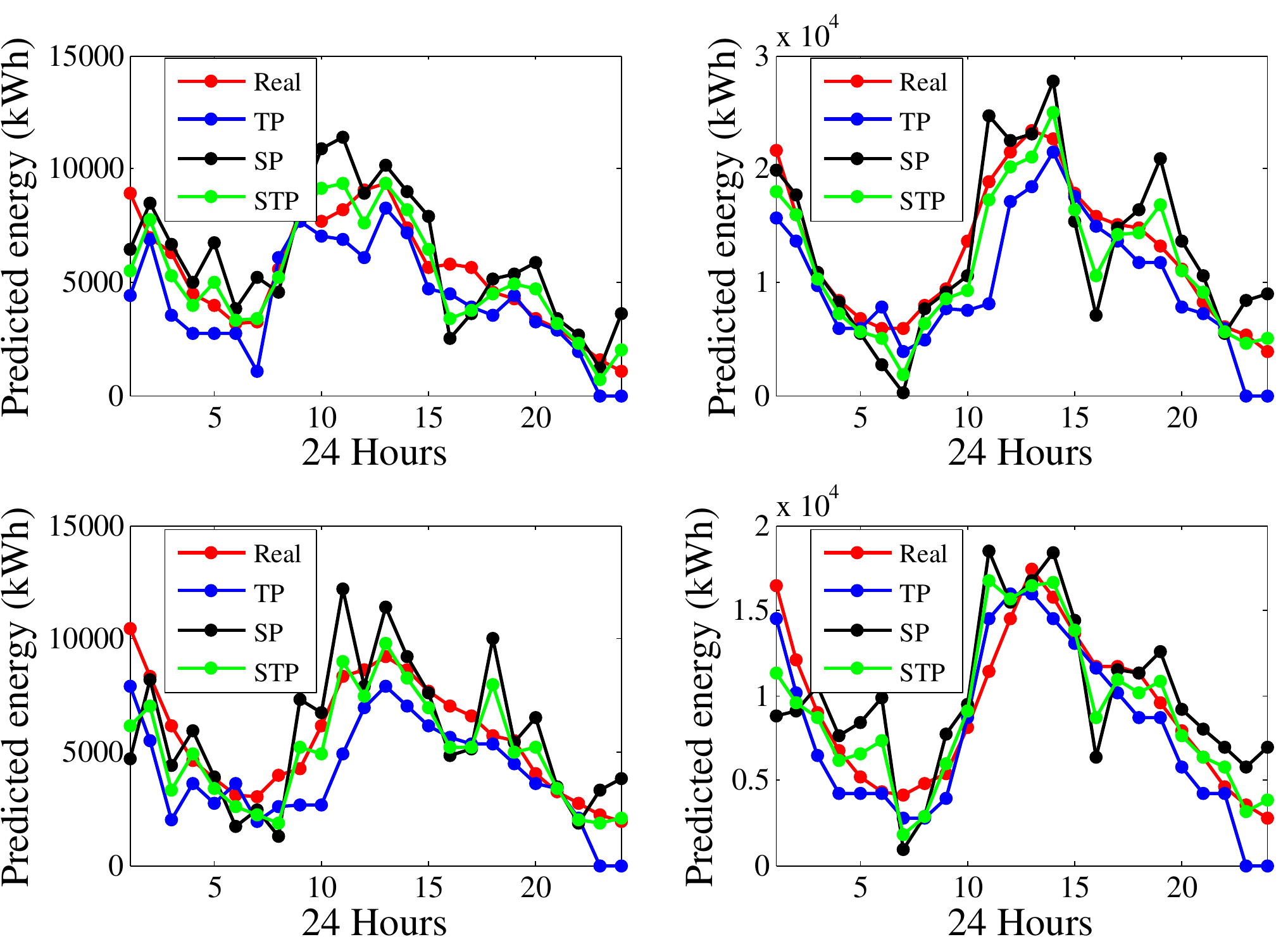}\\
		\caption{Prediction performance of the temporal, spatial and spatio-temporal predictor, compared with the real aggregated energy value of four example regions on a testing day. The prediction results of the temporal predictor and the spatial predictor deviate from the real value in different directions. In most investigated hours, the spatial predictor is more intended to overestimate, while the temporal predictor is more intended to do the opposite. This performance difference between both predictors is further neutralized by the spatio-temporal predictor.}
		\label{fig::temporal_perf}
	\end{center}
\end{figure}
\subsection{Hourly Prediction Performance}
In this experiment, we demonstrate the prediction performance of the temporal, spatial and spatio-temporal predictor for $24$ hours of a testing day in Fig. \ref{fig::temporal_perf}. We select four regions as examples due to space constraint.

Fig. \ref{fig::temporal_perf} shows that the combination of spatial predictor and temporal predictor neutralizes the prediction errors from both. Although the spatio-temporal predictor does not generate better prediction results in all investigated hours, it does achieve a better prediction performance in general. Additionally, it is also noticeable that the superiority of the spatio-temporal predictor remains consistent in all regions. Since the same combination coefficient $\lambda$ is utilized by all regions, this implies that an universal $\lambda$ is applicable for all regions.

\subsection{Average Prediction Performance for All Regions}
\begin{figure}[t]
	\begin{center}
		\includegraphics[scale=0.40]{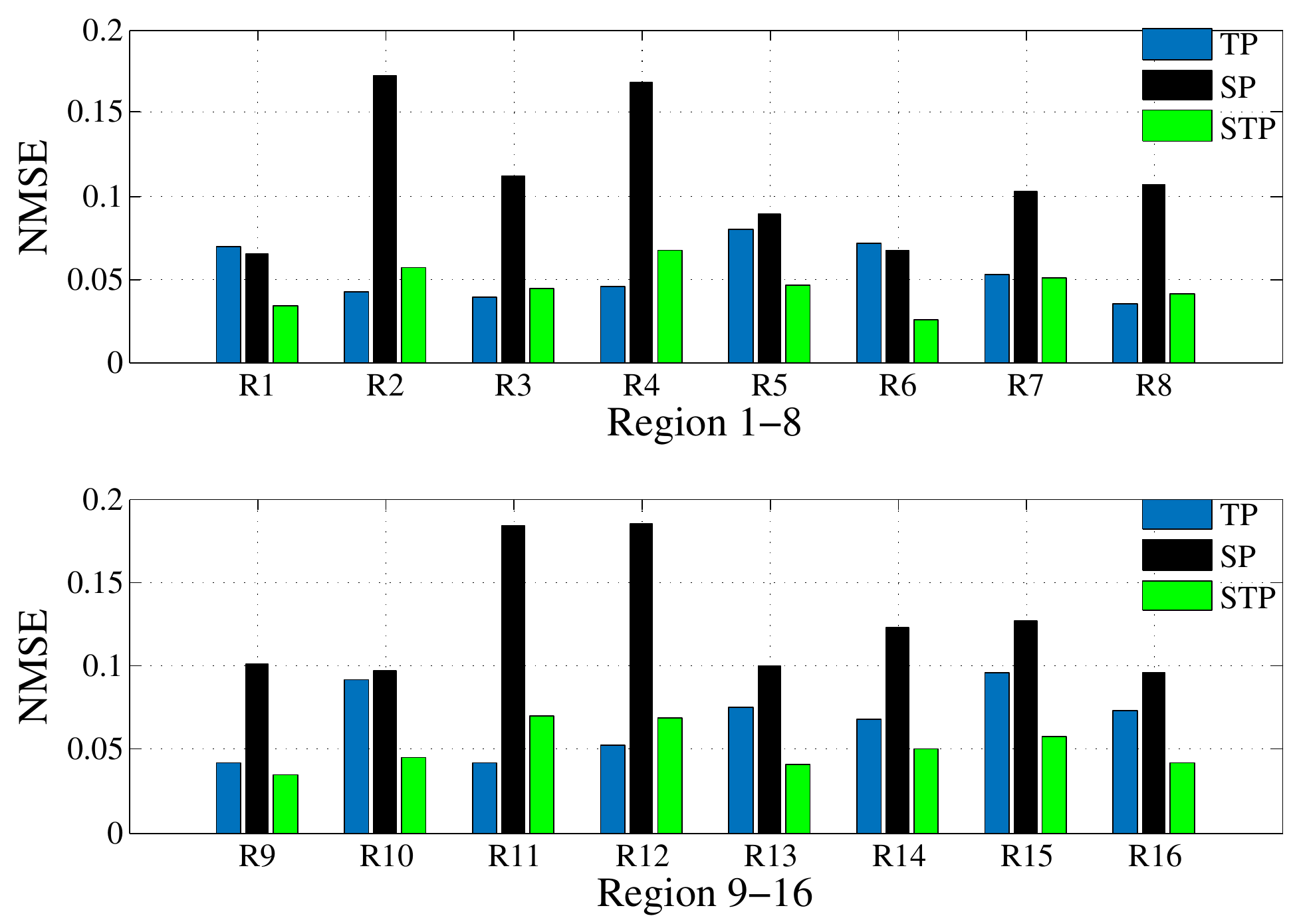}\\
		\caption{The average prediction performance over $24$ hours for all regions. This results shows that the spatial predictor gets worse prediction results in most regions. However, it is also noticeable that the performance of the spatio-temporal predictor is improved in most regions. This is due to the cases when the spatial predictor generates more accurate results compared with the temporal predictor.}
		\label{fig::spatial_perf}
		\vspace{-1.5em}
	\end{center}
\end{figure}
\begin{figure}[b]
	\begin{center}
		\includegraphics[scale=0.40]{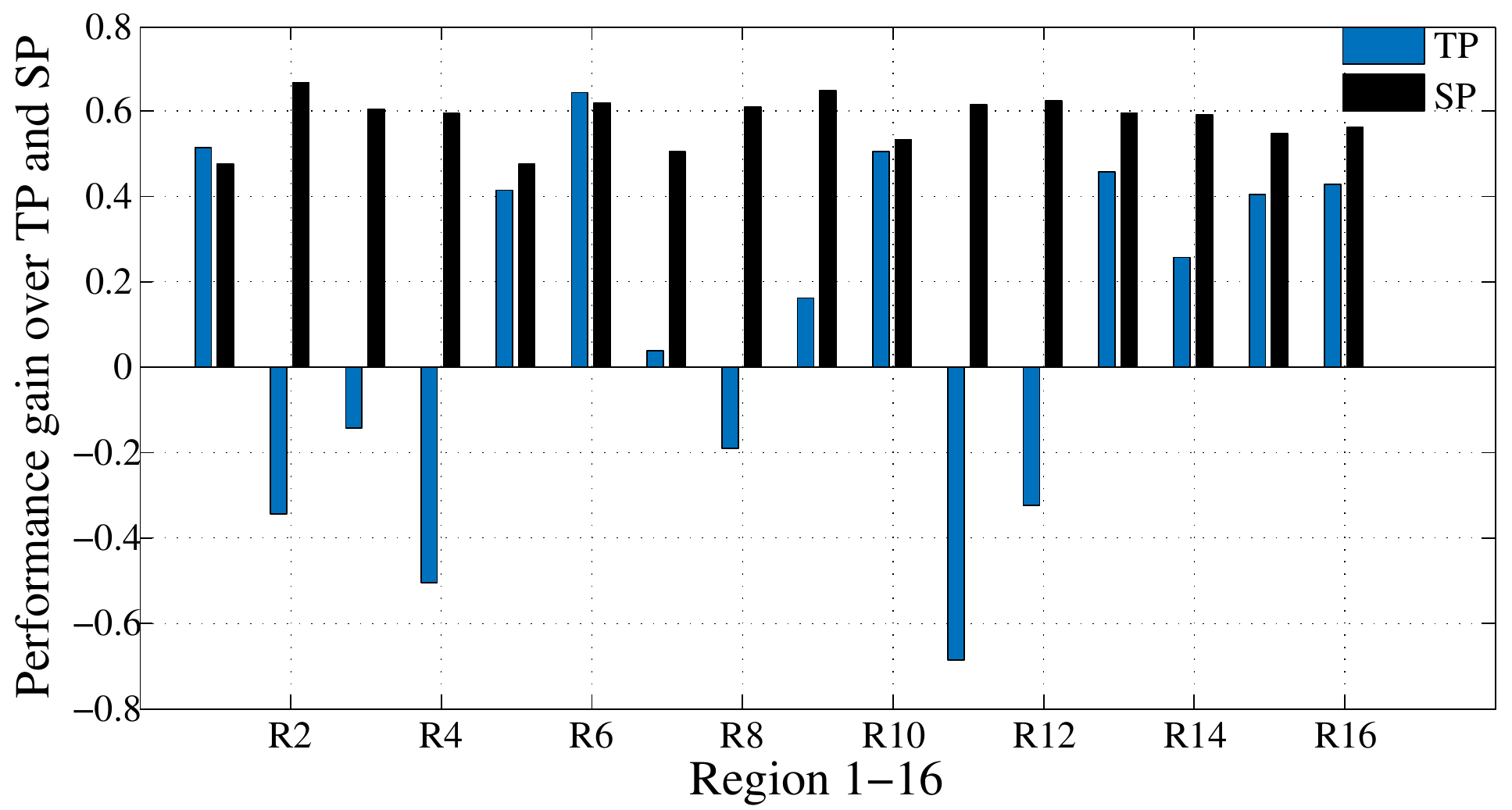}\\
		\caption{The performance gain of the spatio-temporal predictor. In most regions, the spatio-temporal predictor shows an improvement of performance, especially when compared with the spatial predictor. For regions where the spatio-temporal predictor generates worse prediction results compared with the temporal predictor, larger combination weights should be assigned to temporal predictor.}
		\label{fig::perf_gain}
		\vspace{-1.5em}
	\end{center}
\end{figure}
In this experiment, we compare the average prediction performance of the temporal, spatial and spatio-temporal predictor over $24$ hours of all sixteen regions. As shown in Fig. \ref{fig::spatial_perf}, the temporal predictor maintains an average prediction NMSE below $0.1$ in all sixteen regions. This indicates that temporal predictor is robust to spatial changes, as it only focuses on temporal changes within each region. Meanwhile, the spatial predictor shows a general inferiority compared with the temporal predictor. This causes the spatio-temporal predictor to have larger prediction NMSE compared with the temporal predictor for some regions. However, as shown in Fig. \ref{fig::temporal_perf}, this inferiority of the spatial predictor does not hold constantly. Since the spatio-temporal predictor is a weighted combination of both temporal predictor and spatial predictor, it inherits the robustness property from the temporal predictor, and has both temporal predictor and spatial predictor compensate each other's prediction errors. Therefore, the spatio-temporal predictor also maintains an average prediction NMSE below $0.1$ in all sixteen regions. We further show the performance gain of the spatio-temporal predictor over temporal predictor and spatial predictor respectively in Fig. \ref{fig::perf_gain}.

\subsection{Long Term Prediction Performance}
\begin{figure}[t]
	\begin{center}
		\includegraphics[scale=0.46]{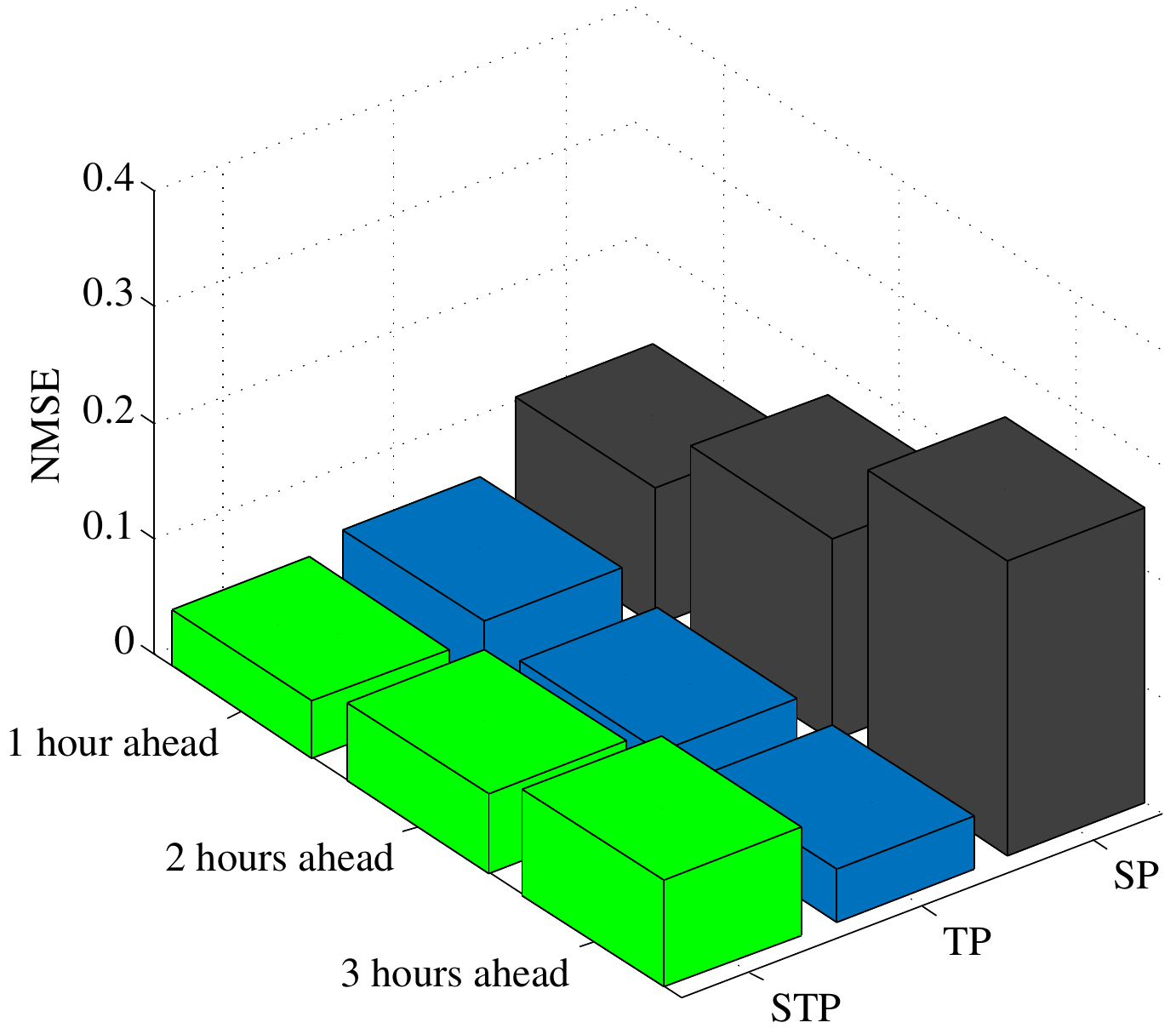}\\
		\caption{Comparison of the NMSE of spatial, temporal and spatio-temporal prediction over three different intervals. The largest NMSE of the spatial predictor ($0.25$) and the spatio-temporal predictor ($0.09$) are generated by $3$ hours ahead predictions. The temporal predictor has the largest NMSE (0.06) for $1$ hour ahead prediction.}
		\label{fig::hour_ahead}
	\end{center}
\end{figure}
Although hourly prediction has been employed by the energy market decades ago, energy prediction which can provide a forecast with longer interval also has traits when the scheduling requires long term information.

In this experiment, we explore the ability of the spatio-temporal predictor to perform two longer term (i.e. two-hour ahead and three-hour ahead) predictions. As shown in Fig. \ref{fig::hour_ahead}, as the prediction interval increases, the prediction performance of three predictors changes differently. Both the spatial predictor and the spatio-temporal predictor show performance degradation (the NMSE results obtained by both predictors increase), especially for the spatial predictor. In term of spatial features, the correlation between current observation and future results fades as the time interval increases. Since the spatial predictor only utilizes spatial features, its prediction performance is influenced more. This influence is further inherited by the spatio-temporal predictor. Meanwhile, the prediction performance of the temporal predictor even shows a slight improvement. This indicates that the linear-chain CRF based temporal predictor is capable of performing longer term prediction. This property prevents the performance of spatio-temporal predictor from degrading more severely.

\subsection{Visualization of Predicted Energy}
\begin{figure}[t]
	\begin{center}
		\includegraphics[scale=0.34]{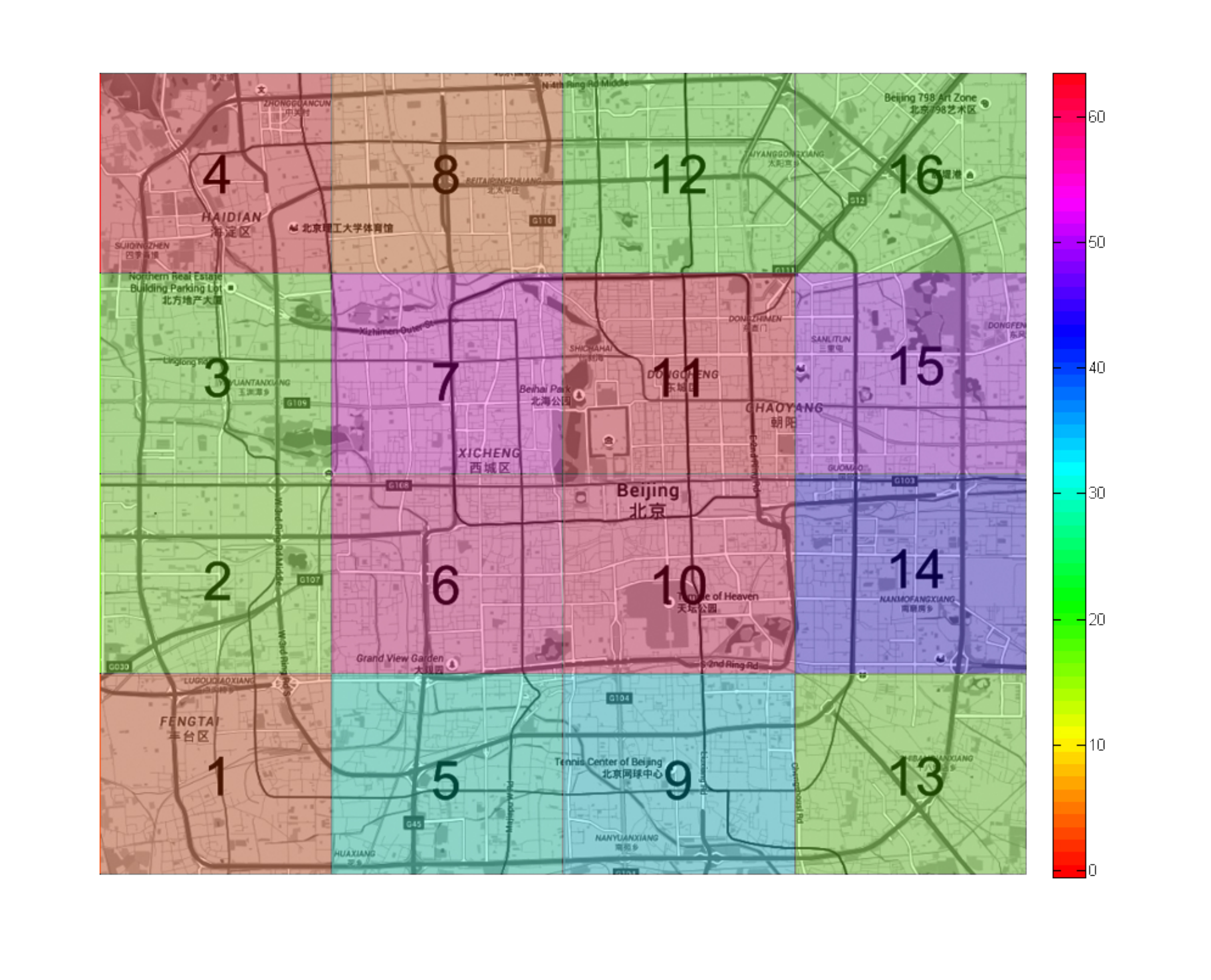}\\
		\caption{Visualization of predicted aggregated energy of each region within Beijing $4$th Ring Road.}
		\label{fig::energy_visual}
	\end{center}
\end{figure}

We provide a realtime visualization of the aggregated energy for practical use. A realtime visualization can assist various aspects of the interactions between the smart gird and aggregated EVs. We utilize the spatio-temporal predictor to predict the aggregated energy of all divided regions within Beijing $4$th Ring Road at 12 p.m. Feb. 8, 2008, as shown in Fig. \ref{fig::energy_visual}. The color mosaic covering the Beijing map indicates different amount of aggregated energy of each region. According to Fig. \ref{fig::energy_visual}, the regions within $3$rd Ring Road of Beijing have more aggregated energy, compared with the rest regions. This is reasonable in that the traffic within $3$rd Ring Road is commonly more crowded, and most EVs still have sufficient stored energy by the time of 12 p.m.

\section{Conclusion and Future Work}\label{sec:conclusion}
In this paper, we focus on exploring spatial and temporal pattern of the aggregated energy carried by a large number of EVs. A spatio-temporal combined prediction framework has been developed to provide predictions of future aggregated energy of multiple city-scale regions. We have extracted both temporal and spatial features from real recorded trajectories data of taxis within Beijing, and studied their influences on the prediction performance. The prediction performance has been demonstrated from both a temporal and spatial perspective. We have shown that the proposed spatio-temporal predictor can maintain a prediction NMSE below $0.1$ in all investigated regions, and has an improved performance compared with both the temporal predictor and the spatial predictor. Furthermore, we have demonstrated that the spatio-temporal predictor is capable of providing longer term predictions with low NMSE.

Our future works will utilize the proposed framework in following aspects, to improve the interaction between EVs and the smart grid:
\begin{enumerate}
	\item Utilizing energy pattern prediction for establishing better interactions between EVs and the smart grid. We plan to utilize the predicted energy pattern to improve the operating efficiency of the smart grid. Since the aggregated energy can be predicted, the smart grid can perform more flexible scheduling.
	\item Utilizing energy pattern prediction to benefit EV charging service providers. More economic efficiency can be brought to EV charging service providers, given hourly prediction of the aggregated energy of their residing areas is available.
	\item Utilizing energy pattern prediction to improve the quality of EV charging service. Predicted energy pattern can help EV charging service providers to schedule and purchase energy in advance, thus providing better service to EV customers.
\end{enumerate}

%

%
%


\end{document}